\documentclass[conference,10pt]{IEEEtran}
\usepackage{mathptmx}
\usepackage[letterpaper,margin=0.75in]{geometry}
\usepackage{amsmath,graphicx}
\usepackage[ruled,vlined]{algorithm2e}
\usepackage{lineno}
\usepackage{amssymb}
\usepackage{soul,color,booktabs,cite}
\usepackage{fancyhdr}

\def\ps@IEEEtitlepagestyle{%
  \def\@oddfoot{\mycopyrightnotice}%
  \def\@evenfoot{}%
}
\def\mycopyrightnotice{%
  {\footnotesize The copyright belongs to me!\hfill}
  \gdef\mycopyrightnotice{}
}

\newcommand{\EQ}{\begin{equation}}
\newcommand{\NQ}{\end{equation}}
\newcommand{\ER}{\begin{eqnarray}}
\newcommand{\NR}{\end{eqnarray}}
\newcommand{\ERS}{\begin{eqnarray*}}
\newcommand{\NRS}{\end{eqnarray*}}
\newcommand{\bit}{\begin{itemize}}
\newcommand{\ben}{\begin{enumerate}}
\newcommand{\eben}{\end{enumerate}}
\newcommand{\ebit}{\end{itemize}}
\newcommand{\bzero}{{\bf 0}}
\newcommand{\ba}{{\bf a}}

\newcommand{\bbf}{{\bf f}}

\newcommand{\bx}{{\bf x}}

\newcommand{\bA}{{\bf A}}

\newcommand{\bC}{{\bf C}}

\newcommand{\bU}{{\bf U}}

\ifCLASSINFOpdf
\else
\fi
\begin{document}
%
\title{Zero-shot Image Recognition Using Relational Matching, Adaptation and Calibration}
%
%
%

\author{Debasmit~Das~\IEEEmembership{Student Member,~IEEE,}
        and~C.S. George~Lee~\IEEEmembership{Life~Fellow,~IEEE}
}

%
%

\markboth{Journal of \LaTeX\ Class Files,~Vol.~14, No.~8, August~2015}%
{Shell \MakeLowercase{\textit{et al.}}: Bare Demo of IEEEtran.cls for IEEE Journals}
%




\maketitle

\begin{abstract}
Zero-shot learning (ZSL) for image classification focuses on
recognizing novel categories that have no labeled data available for training. 
The learning is generally carried out with the help of 
mid-level semantic descriptors associated with each class. 
This semantic-descriptor space is generally shared by 
both seen and unseen categories. 
However, ZSL suffers from hubness, domain discrepancy 
and biased-ness towards seen classes. 
To tackle these problems, we propose a three-step approach to zero-shot learning. 
Firstly, a mapping is learned from the semantic-descriptor space to the image-feature space. 
This mapping learns to minimize both one-to-one and pairwise distances 
between semantic embeddings and the image features of the corresponding classes. 
Secondly, we propose test-time domain adaptation 
to adapt the semantic embedding of the unseen classes to the test data. 
This is achieved by finding correspondences between the semantic descriptors 
and the image features. 
Thirdly, we propose scaled calibration on the classification scores of the seen classes. 
This is necessary because the ZSL model is biased towards 
seen classes as the unseen classes are not used in the training. 
Finally, to validate the proposed three-step approach, 
we performed experiments on four benchmark datasets 
where the proposed method outperformed previous results. 
We also studied and analyzed the performance of 
each component of our proposed ZSL framework.
\end{abstract}

\footnotetext{Debasmit Das and C.S. George Lee are with the School
of Electrical and Computer Engineering, Purdue University, West Lafayette,
IN, 47907 USA. E-mail: \{dsdas, csglee\}@purdue.edu.}
\footnotetext[2]{This work was supported in part by 
the National Science Foundation under Grant IIS-1813935. 
Any opinion, findings, and conclusions or recommendations 
expressed in this material are those of the authors and do not necessarily 
reflect the views of the National Science Foundation.}
\footnotetext[2]{We also gratefully acknowledge the support of NVIDIA Corporation 
for the donation of a TITAN XP GPU used for this research.}


%
\IEEEpeerreviewmaketitle

\section{Introduction}
%
%
%
%
\IEEEPARstart{R}{ecent} work on visual recognition focuses on 
the importance of obtaining large labeled datasets 
such as ImageNet~\cite{deng2009imagenet}. 
Large-scale datasets when used for 
training deep neural network models tend to produce 
state-of-the-art results on visual recognition~\cite{krizhevsky2012imagenet}. 
However, in some cases, it may be difficult to obtain a large number of samples 
for certain rare or fine-grained categories. 
Hence, recognizing these rare categories become difficult. 
Humans, on the other hand, can easily recognize these rare categories 
by identifying the semantic description of the new category 
and how it is related to the seen categories. 
For example, a person can identify a new animal zebra by 
identifying the semantic description of a zebra having 
black and white stripes and looking like a horse.
A similar approach is undertaken for  
learning models to recognize unseen and rare categories. 
This learning scenario is known as zero-shot learning (ZSL) 
because zero labeled samples of the unseen categories 
are available for the training stage. ZSL has promising ramifications in autonomous vehicles, medical imaging, robotics, etc., where it is difficult to annotate images of novel categories but high-level semantic descriptions of classes can be obtained easily.

\begin{figure}[t]
\centering
\includegraphics[width=8cm]{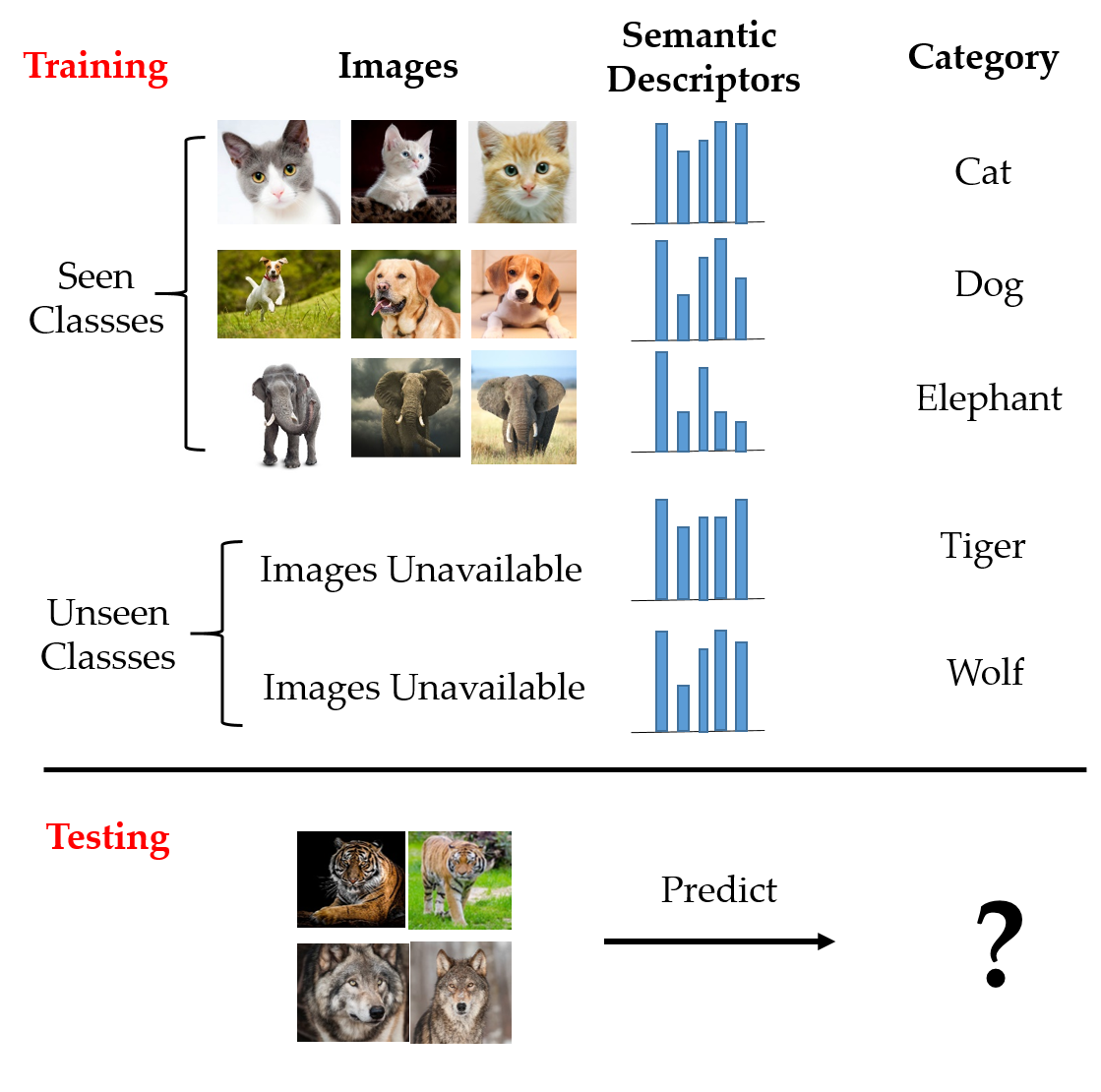}
\vspace*{-0.1in}
\caption{Depiction of the zero-shot learning problem. 
During training, we have lots of labeled images from seen classes (cat, dog, elephant) 
but no labeled images from unseen classes. 
We do have semantic descriptors of all the classes available. 
Using all the information, the goal is to recognize the unseen classes.}
\label{fig:zsl}
\vspace*{-0.2in}
\end{figure}

To be able to recognize unseen categories, 
we usually train a learning model using a large collection of labeled samples 
from the seen categories and then adapt it to unseen categories. 
For zero-shot recognition, the seen and the unseen categories 
are related through a high-dimensional vector space 
known as semantic-descriptor space. 
Each category is assigned a unique semantic descriptor. 
Examples of semantic descriptor can be manually 
defined attributes~\cite{lampert2014attribute} 
or automatically extracted word vectors~\cite{word2vec}. 
Figure~\ref{fig:zsl} depicts the ZSL problem in terms of 
how much information is available during training and testing.

Most ZSL methods involve mapping from the visual feature space 
to the semantic-descriptor space or 
vice versa~\cite{Zhang_2017_CVPR,akata2016label,frome2013devise,socher2013zero}. 
Sometimes, both the visual features and the semantic descriptors 
are mapped to a common feature space~\cite{zhang2016zero,changpinyo2016synthesized}. 
Most of these mapping-based approaches 
learn an embedding function for samples and semantic descriptors. 
The embedding is learned by minimizing a similarity function 
between the embedded samples and the corresponding embedded semantic descriptors. 
Thus, most ZSL methods differ in the choice of the embedding and similarity functions. 
Lampert et. al~\cite{lampert2014attribute} used linear classifiers, 
identity function and Euclidean distance for the sample embedding, 
semantic embedding and similarity metric, respectively. 
Romera-Paredes et al.~\cite{romera2015embarrassingly} 
used linear projection, identity function and dot product. 
ALE~\cite{akata2016label}, DEVISE~\cite{frome2013devise}, 
SJE~\cite{akata2015evaluation} all used a bilinear compatibility framework, 
where the projection was linear and the similarity metric was a dot product. 
They used different variations of pairwise ranking objective to train the model. 
LATEM~\cite{xian2016latent} was an extension of the above method, 
which used piecewise linear projections to account for the non-linearity. 
CMT~\cite{socher2013zero} used a neural network to map image features 
to semantic descriptors with an additional novelty detection stage to detect unseen categories.
SAE~\cite{kodirov2017semantic} used an auto-encoder-based approach, 
where the image feature is linearly mapped to a semantic descriptor 
as well as being reconstructed from the semantic-descriptor space. 
DEM~\cite{Zhang_2017_CVPR} used a neural network to map 
from a semantic-descriptor space to an image-feature space. 
 
After the embedding is carried out, 
classification is performed using the nearest-neighbor search. 
An earlier study~\cite{radovanovic2010hubs} showed that the nearest-neighbor search 
in such a high-dimensional space suffers from the \emph{hubness phenomenon} 
because only a certain number of data-points becomes nearest neighbor 
or hubs for almost all the query points, resulting in erroneous classification results. 
However, Shigeto et al.~\cite{shigeto2015ridge} showed that 
mapping from a semantic-descriptor space 
to a visual-feature space does not aggravate the hubness problem. 
Thus, in this paper, we pursue a semantic-descriptor-space
to a visual-feature-space mapping approach. 
We further introduce the concept of relative features 
that uses pairwise relations between data-points. 
This not only provides additional structural information about the data 
but also reduces the dimensionality of the feature space implicitly~\cite{2stage}, 
thus alleviating the hubness problem.

Zero-shot learning further suffers from a projection-domain-shift problem 
because the mapping from the semantic-descriptor space 
to the visual-feature space is learned from 
the data belonging to only the seen categories. 
As a result, the projected semantic descriptors of 
the unseen categories are misplaced from the unseen test-data distribution.  
Fu et al.~\cite{fu2015transductive} identified the domain-shift problem 
and used multiple semantic information sources and label propagation 
on unlabeled data from the unseen categories to counter the problem. 
Kodirov et al.~\cite{kodirov2015unsupervised} cast ZSL 
as a dictionary-learning problem and constrained the dictionary 
of the seen and unseen data to be close to each other. 
This transductive approach is unrealistic 
as it assumes access to the unlabeled test data from unseen categories 
during the training stage. 
At the very least, we could carry out the test-time post-processing 
of the semantic descriptors. 
For test-time adaptation, we propose to find correspondences 
between the projected semantic descriptors and the unlabeled test data 
after which the descriptors are further mapped to the corresponding data-points. 
This is inspired by recent work on local correspondence-based approach 
to unsupervised domain adaptation~\cite{das2018sample}, 
which produces better results than global domain-adaptation methods.

Another problem with ZSL is that models 
are generally evaluated only on unseen categories. 
In a real-world scenario, we expect the seen categories to appear 
more frequently compared to the unseen categories. 
As a result, it is appropriate to test our model on both seen and unseen categories. 
This evaluation setting is known as Generalized Zero-Shot Learning (GZSL) 
and was initially introduced by Chao, et al.~\cite{chao2016empirical}. 
They found that the performance of unseen categories 
in the GZSL setting was poor and proposed a shifted-calibration mechanism 
to improve the performance. 
This shifted-calibration mechanism lowers the classification scores of the seen categories. 
We propose to develop a scaled-calibration mechanism to study the effect on recognition performance. 
This has an effect of changing 
the effective variance of a class and is therefore more interpretable.

Other methods for ZSL include hybrid and synthesized methods. 
Hybrid models expressed image features or semantic embeddings 
as a combination/mixture of existing seen features or semantic embeddings. 
Semantic Similarity Embedding (SSE)~\cite{zhang2015zero} 
exploits class relationship at both the image-feature and semantic-descriptor spaces 
to map them into a common embedding space. 
Our proposed ZSL method also exploits pairwise relationships 
between classes by minimizing the discrepancy between 
the projected semantic descriptors and the corresponding class prototype 
obtained from the image features. 
CONSE~\cite{norouzi2013zero} learns the probability of a seen sample 
belonging to a seen class and uses the probability of an unseen sample 
belonging to seen classes to relate to the semantic-descriptor space. 
Synthetic Classifiers (SYNC)~\cite{changpinyo2016synthesized} 
learn a mapping between the semantic-embedding space 
and the model-parameter space. 
The model parameters of the classes are represented as a combination of phantom classes, 
the relationship with which is encoded through a weighted bipartite graph. 
Synthesized methods generally convert ZSL into 
a standard supervised-learning problem 
by generating samples for the unseen categories. 
Some of these methods 
include~\cite{verma2018generalized,guo2017synthesizing}. 
The limitations of these methods lie in not being able 
to generate samples very close to the true distribution. 
A more comprehensive overview of recent work on ZSL can be found in~\cite{xian2018zero}.

To summarize, we propose a three-step approach to zero-shot learning. 
Firstly, to prevent aggravating the hubness problem, 
a mapping is learned from the semantic-descriptor space to the image-feature space that minimizes both one-to-one and pairwise distances 
between semantic embeddings and the image features. 
Secondly, to alleviate the domain-shift problem at test time, 
we propose a domain-adaptation method that finds
correspondences between the semantic descriptors and the image features of test data.
Thirdly, to reduce biased-ness in the GZSL setting, 
we propose scaled calibration on the classification scores of the seen classes
to balance the performance on the seen and unseen categories.
Finally, we evaluated our proposed approach on four standard ZSL datasets 
and compared our approach against state-of-the art methods 
followed by further analyzing the contribution of each component of our approach.
\begin{figure*}[t]
\centering
\includegraphics[width=16cm]{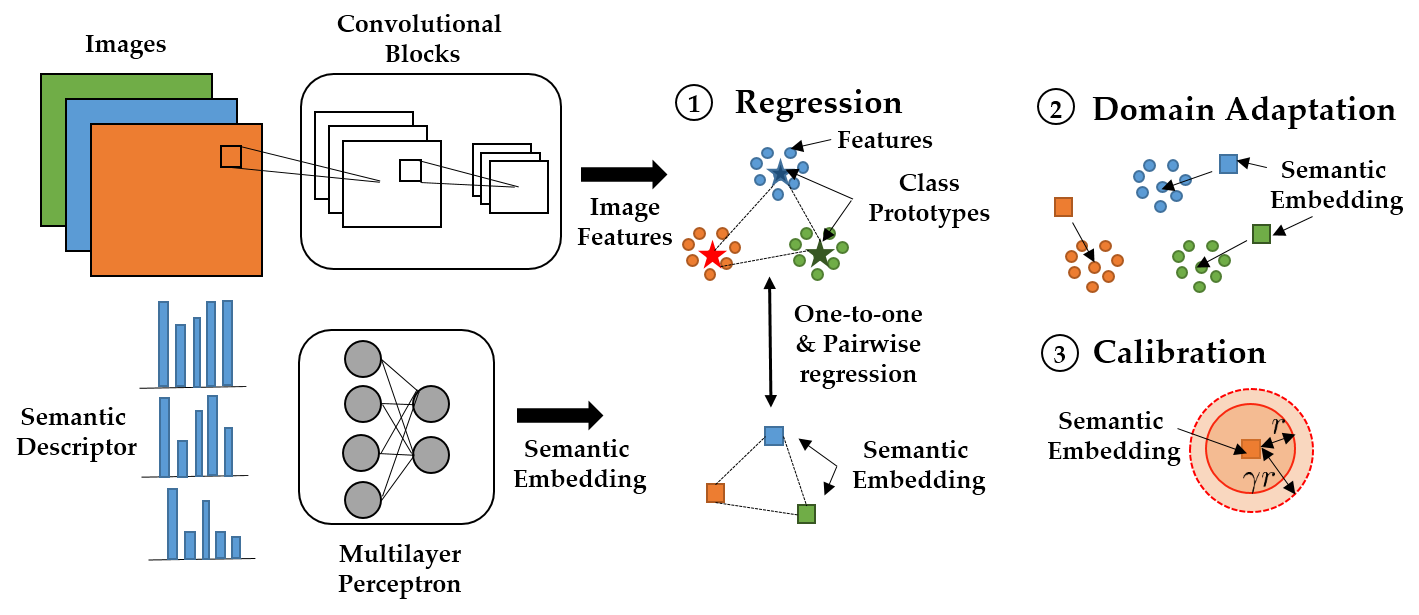}
\vspace*{-0.1in}
\caption{The semantic descriptors are mapped to the image-feature space through the multi-layer perceptron. Then the semantic embeddings are regressed to the corresponding features through one-to-one and pairwise relations. After that, the semantic embeddings of the unseen classes are adapted to the unseen test data. This is followed by scaled calibration during testing when classification scores of seen classes are modified.}
\label{fig:approach}
\vspace*{-0.2in}
\end{figure*}

\section{Methodology}
\subsection{Problem Description}
Let the training dataset $\mathcal{D}_{tr}$ 
consist of $N_{tr}$ samples such that 
$\mathcal{D}_{tr}=\{(\mathbf{x}_i, \mathbf{a}_i, y_i), i=1,2,\ldots,N_{tr}\}$. 
Here, $\mathbf{x}_i \in \mathbb{R}^{m\! \times \! n \! \times \!c}$ 
is an image sample ($m\! \times \!n$ is the image size and $c$ is the number of channels) 
and $\mathbf{a}_i \in \mathbb{R}^{s}$ is the semantic descriptor 
of the sample's class. 
Each semantic descriptor $\mathbf{a}_i$ is uniquely associated with 
a class label $y_i \in \mathcal{Y}_{tr}$. 
The goal of ZSL is to predict the class label $y_j \in \mathcal{Y}_{te}$ 
for the $j^{th}$ test sample $\mathbf{x}_j$. 
In the traditional ZSL setting, 
we assume that $\mathcal{Y}_{tr} \cap \mathcal{Y}_{te} = \varnothing$; 
that is, the seen (training) and the unseen (testing) classes are disjoint. 
However, in the GZSL setting, both seen and unseen classes 
can be used for testing; 
that is, $\mathcal{Y}_{tr} \subset \mathcal{Y}_{te}$. 
In the training stage, we have the semantic descriptors 
of both the seen and unseen classes available 
but no labeled training data of the unseen classes are available. 
The overall framework of our proposed ZSL approach is shown in Fig.~\ref{fig:approach}.

\subsection{Relational Matching}
Our goal is to learn a mapping $\mathbf{f}(\cdot)$ 
that maps a semantic descriptor $\mathbf{a}_i$ 
to its corresponding image feature $\mathbf{\phi}(\mathbf{x}_i)$. 
Here, $\mathbf{x}_i$ is an image and $\mathbf{\phi}(\cdot)$ 
represents a CNN architecture that extracts a high-dimensional feature map. 
The mapping $\bbf(\cdot)$ is a fully-connected neural network. 
Since our goal is to make the embedded semantic descriptor 
close to the corresponding image feature, 
we use a least square loss function to minimize the difference. 
We also need to regularize the parameters of $\mathbf{f}(\cdot)$. 
Including these costs and averaging over all the instances, 
our initial objective function $\mathcal{L}_1$ is as follows:
\begin{equation}
\mathcal{L}_1 = \frac{1}{N_{tr}}\sum_{i=1}^{N_{tr}}||\mathbf{f}(\ba_i)-\mathbf{\phi}(\mathbf{x}_i)||_2^2 + \lambda_rg(\bbf)~,
\end{equation}
where $g(\cdot)$ is the regularization loss for the mapping function. 
The loss function $\mathcal{L}_1$ minimizes the point-to-point discrepancy 
between the semantic descriptors and the image features. 
To account for the structural matching between 
the semantic-descriptor space and the image-feature space, 
we try to minimize the inter-class pairwise relations in these two spaces. 
Thus, we construct relational matrices for both the semantic descriptors and image features. 
The semantic relational matrix $\mathbf{D}_a$ is established 
such that each element, 
$[\mathbf{D}_a]_{uv} = ||\bbf(\ba^u)-\bbf(\ba^v)||_2^2$, 
where $\ba^u$ and $\ba^v$ are semantic descriptors 
of seen categories $u$ and $v$, respectively. 
The image feature relational matrix $\mathbf{D}_\phi$ 
is constructed such that each element, 
$[\mathbf{D}_\phi]_{uv} = ||\overline{\mathbf{\phi}}^u-\overline{\mathbf{\phi}}^v||_2^2$, 
where $\overline{\mathbf{\phi}}^u$ and $\overline{\mathbf{\phi}}^v$ 
are mean representations of the categories $u$ and $v$, respectively. 
$\overline{\mathbf{\phi}}^u$ can be represented as 
\begin{equation}
\overline{\mathbf{\phi}}^u=\frac{1}{|\mathcal{Y}^{u}_{tr}|}\sum_{y_i \in \mathcal{Y}^{u}_{tr}}^{}\mathbf{\phi}(\bx_i)~,
\end{equation}
where the summation is over the representations of class $u$, 
and $|\mathcal{Y}^{u}_{tr}|$ is the cardinality of the training set of class $u$. 
A similar formula holds for class $v$. 
For structural alignment, we want the two relational matrices, 
$\mathbf{D}_a$ and $\mathbf{D}_{\phi}$, 
to be close to one another. 
Hence, we want to minimize the structural alignment loss function $\mathcal{L}_2$,
\begin{equation}
\mathcal{L}_2=||\mathbf{D}_a-\mathbf{D}_{\phi}||_F^2~,
\end{equation}
where $||\cdot||_F^2$ stands for the Frobenius norm. 
Combining the loss functions $\mathcal{L}_1$ and $\mathcal{L}_2$, 
we have the total loss $\mathcal{L}_{total}$,
\begin{equation}
\mathcal{L}_{total} = \mathcal{L}_1 + \rho \mathcal{L}_2~,
\end{equation}
where $\rho \geq 0$ weighs the loss contribution of $\mathcal{L}_2$.
$\mathcal{L}_{total}$ is to be optimized with respect to 
the parameters of the semantic-descriptor-to-visual-feature-space mapping $\bbf(\cdot)$.

\subsection{Domain Adaptation}
After the training is carried out, 
domain discrepancy may be present between 
the mapped semantic descriptors and the image features of unseen categories. 
This is because the unseen data has not been used in the training 
and our regularized model does not 
generalize well for the unseen categories. 
Hence, we need to adapt the mapped semantic descriptors 
for the unseen categories using the test data from the unseen categories. 
Let the mapped descriptors for the unseen categories be stacked vertically 
in the form of a matrix $\bA \in \mathbb{R}^{n_u\!\! \times \!\!d}$, 
where $n_u$ is the number of unseen categories 
and $d$ is the dimension of the mapped semantic-descriptor space, 
and therefore it is also the dimension of the image-feature space. 
Let $\bU \in \mathbb{R}^{o_u\!\! \times \!\!d}$ be the unseen test dataset, 
where $o_u$ is the number of test instances from the unseen categories. 
For adapting the mapped descriptors, 
we propose to find the point-to-point correspondence 
between the descriptors and the test data. 
Let the correspondence be represented as a matrix 
$\bC \in \mathbb{R}^{n_u\! \times \!o_u}$. 
We want to rearrange the rows of $\bU$ such that 
each row of the modified matrix corresponds to the row in $\bA$. 
This is done by minimizing the following loss function $\mathcal{L}_3$,
\begin{equation}
\mathcal{L}_3=||\bC\bU-\bA||_F^2~.
\end{equation}

This loss function enforces that $\bC\bU$ produces the adapted semantic descriptors. 
However, a problem may exist that an instance in $\bU$ 
corresponds to more than one descriptor in $\bA$. 
This would essentially result in a test sample corresponding to more than one category. 
To avoid that, we use an additional group-based regularization function 
$\mathcal{L}_4$ using Group-Lasso,
\begin{equation}
\mathcal{L}_4=\sum_{j}\sum_{c}||[\bC]_{I_cj}||_2~,
\end{equation}
where $I_c$ corresponds to the indices of those rows in $\bA$ 
that belong to the unseen class $c$. 
Therefore, $[\bC]_{I_cj}$ is the vector consisting of 
the row indices from $I_c$ and the $j^{th}$ column. 
Since $\bC$ is a correspondence matrix, 
some constraints should be enforced such as $\bC \geq \mathbf{0}$, 
$\bC\mathbf{1}_{o_u}=\mathbf{1}_{n_u}$ 
and $\bC^{T}\mathbf{1}_{n_u}=\frac{n_u}{o_u}\mathbf{1}_{o_u}$, 
where $\mathbf{1}_{n}$ is an $n\! \times \!1$ vector of one's. 
The second equality constraint is scaled by the factor $\frac{n_u}{o_u}$ 
to account for the difference in the number of instances 
in the mapped semantic-descriptor space and the image-feature space 
for the unseen categories. 
Hence, the domain adaptation optimization problem becomes 
\begin{equation}
\underset{\bC}{\text{min}} \hspace{0.1em} \left\{\mathcal{L}_3 + \lambda_g \mathcal{L}_4\right\}  \hspace{0.5em} s.t. \hspace{0.5em} \bC \geq \mathbf{0}, \bC\mathbf{1}_{o_u}=\mathbf{1}_{n_u},\bC^{T}\mathbf{1}_{n_u}=\frac{n_u}{o_u}\mathbf{1}_{o_u}~,
\label{eq7}
\end{equation}
where $\lambda_g$ weighs the loss function $\mathcal{L}_4$.

The above optimization problem is convex and can be efficiently solved 
using the conditional gradient method~\cite{frank1956algorithm}. 
The conditional gradient method requires solving 
a linear program as an intermediate step over the constraints 
$\bC \in \mathcal{D}=\{\bC: \bC \geq \bzero, \bC\mathbf{1}_{o_u}=\mathbf{1}_{n_u}, 
\bC^{T}\mathbf{1}_{n_u}=\frac{n_u}{o_u}\mathbf{1}_{o_u}\}$ 
as shown in Algorithm 1. 
The linear program of finding the intermediate variable 
$\bC_d$ in Algorithm 1 can be easily solved 
using a network simplex formulation 
of the earth-mover's distance problem~\cite{bonneel2011displacement}.
\begin{algorithm}[]
\SetAlgoLined
 \textbf{Intitialize :} $\bC_0 = \frac{1}{(n_uo_u)}\mathbf{1}_{n_u\! \times \!o_u}$, $t=1$\\
 \textbf{Repeat}\\
 \quad $\bC_d = \underset{\bC}{\text{argmin}} \hspace{0.5em} \text{Tr}(\nabla_{\bC=\bC_0} (\mathcal{L}_3 + \lambda_g \mathcal{L}_4)^{T}\bC),  \hspace{0.5em} s.t.  \hspace{0.5em} \bC \in \mathcal{D}$ \\
 \quad $\bC_1 = \bC_0 + \alpha(\bC_d - \bC_0), \quad \text{for} \quad \alpha = \frac{2}{t+2}$ \\
 \quad $\bC_0=\bC_1$ \quad \text{and} \quad $t=t+1$ \\
 \textbf{Until} Convergence \\
 \textbf{Output :} $\bC_0 = \text{arg}\min \limits_{\bC}\{\mathcal{L}_3 + \lambda_g \mathcal{L}_4\} \quad s.t. \quad \bC \in \mathcal{D}$
\caption{Conditional Gradient Method (CG)}
\end{algorithm}

Once the final solution of the correspondence matrix $\bC_0$ 
in Algorithm 1 is obtained, 
we inspect $\bC_0$. 
For each test instance, 
we assign the class correspondence to the highest value of the correspondence variable. 
This is done for all the test instances. 
The new semantic descriptors are obtained by taking the mean 
of the feature instances belonging to the corresponding class. 
The adapted semantic descriptors are then stacked vertically in the matrix $\bA'$.

\subsection{Scaled Calibration}
In the GZSL setting, 
it is known that the classification results are biased 
towards the seen categories~\cite{chao2016empirical}. 
To counteract the bias, 
we propose the use of multiplicative calibration on the classification scores. 
In our case, we use 1-Nearest Neighbor (1-NN) 
with the Euclidean distance metric as the classifier. 
The classification score for a test point 
is given by the Euclidean distance of the test image feature 
to the mapped semantic descriptor of a category. 
For a test point $\bx$, 
we adjust the classification scores on the seen categories as follows
\begin{equation}
\hat{y}=\underset{c \in \mathcal{T}}{\text{argmin}}
\hspace{0.5em}||\bx-\bbf(\ba^c)||_2\cdot\mathbb{I}[c \in \mathcal{S}]~,
\end{equation} 
where $\mathbb{I}[\cdot]=\gamma$ if $c \in \mathcal{S}$ 
and 1 if $c \in \mathcal{U}$ and $\mathcal{S} \cup \mathcal{U} = \mathcal{T}$. 
Here, $\mathcal{S}, \mathcal{U}$ and $\mathcal{T}$ 
represent the sets of seen, unseen and all categories, respectively. 
The effect of scaling is to change the effective variance of the seen categories. 
When the nearest-neighbor classification is carried out 
with the Euclidean distance metric, 
it assumes that all classes have equal variance. 
But since the unseen categories are not used 
for learning the embedding space, 
the variance of the unseen-category features is not accounted for. 
That is why the Euclidean distance metric 
for the seen categories needs to be adjusted for. 
For $\gamma > 1$, if we obtain a balanced performance 
between the seen and unseen classes, 
it implies that the variance of the seen classes has been overestimated. 
Similarly, if we obtain a balanced performance for $\gamma < 1$, 
it means that the variance of the seen classes has been underestimated. 
The overall procedure of our proposed zero-shot learning method 
from training to testing is given in Algorithm 2.
\begin{algorithm}[]
\SetAlgoLined
\textbf{Input:} Training Dataset $\{(\mathbf{x}_i, \mathbf{a}_i, y_i)\}_{i=1}^{N_{tr}}$\\
\textbf{Parameters:} $\lambda_r$, $\rho$, $\lambda_g$, $\gamma$\\
\textbf{Repeat} (Training)\\
\quad Sample Minibatch of $\{(\bx_i,\ba_i)\}$ pairs\\
\quad Gradient descent $\mathcal{L}_1 + \rho \mathcal{L}_2$ w.r.t parameters of $\bbf(\cdot)$\\
\textbf{Until} Convergence \\
\textbf{Input:} Test Dataset $\{(\mathbf{x}_i)\}_{i=1}^{N_{te}}$\\
\quad Apply Algorithm 1 to obtain adapted descriptors \\
\quad of unseen classes $\bA'$ (Adaptation)\\
\textbf{Repeat} for each test point $\bx$ (Testing)\\
\quad $\hat{y}=\underset{c \in \mathcal{T}}{\text{argmin}}\hspace{0.5em}||\bx-\bbf(\ba^c)||_2\cdot\mathbb{I}[c \in \mathcal{S}]$ (Calibration)\\
\textbf{Until} all test points covered
\caption{Proposed Zero-shot Learning Algorithm}
\end{algorithm}

\section{Experimental Results}
Following the previous experimental settings~\cite{xian2018zero}, 
we used the following four datasets for evaluation: 
\textbf{AwA2}~\cite{lampert2014attribute} (Animal with Attributes) 
contains 37,322 images of 50 classes of animals. 
40 classes of animals are considered to be the seen categories 
while 10 classes of animals are considered to be the unseen categories. 
Each class is associated with a 85-dimensional continuous semantic descriptor. 
\textbf{aPY}~\cite{farhadi2009describing} (attribute Pascal and Yahoo) 
consists of 20 seen categories and 12 unseen categories. 
Each category has an associated 64-dimensional semantic descriptor. 
\textbf{CUB}~\cite{welinder2010caltech} (Caltech-UCSD Birds-200-2011) 
is a fine-grained dataset consisting of 11,788 images of birds. 
For evaluation, all the bird categories are split into 150 seen classes and 50 unseen classes. 
Each class is associated with a 312-dimensional continuous semantic descriptor. 
\textbf{SUN}~\cite{patterson2012sun} (Scene UNderstanding database) 
consists of 14340 scene images. 
Among these, 645 scene categories are selected as seen categories 
while 72 categories are selected as unseen categories 
and it consists of a 102-dimensional semantic descriptor.

For the purpose of evaluation, 
we used class-wise accuracy because it prevents dense-sampled classes 
from dominating the performance. 
Accordingly, class-wise accuracy is averaged as follows
\begin{equation}
acc=\frac{1}{|\mathcal{Y}|}\sum_{y=1}^{|\mathcal{Y}|}\frac{\text{No. of correct predictions in class}\hspace{0.5em} y}{\text{No. of samples in class}\hspace{0.5em} y},
\end{equation}
where $|\mathcal{Y}|$ is the number of testing classes.
In the GZSL case, class-wise accuracy of both seen and unseen classes 
are obtained separately and then averaged using harmonic mean $H$~\cite{xian2018zero}. 
This is done so that the performance on seen classes 
does not dominate the overall accuracy,
\begin{equation}
H=\frac{2\times acc_s\! \times \!acc_u}{acc_s + acc_u},
\end{equation}
where $acc_s$ and $acc_u$ are the class-wise accuracy 
on seen and unseen categories, respectively. 
In the GZSL classification setting, 
the search space of predicted categories consists of both seen and unseen categories. Based on \cite{xian2018zero} and for fair comparison, a single trial of experimental results on a large batch of training and testing dataset is reported.

For the experiments, 
we used a two-layer feedforward neural network 
for the semantic embedding $\bbf(\cdot)$. 
The dimensionality of the hidden layer was chosen as 1600, 1600, 1200 
and 1600 for the \textbf{AwA2}, \textbf{aPY}, \textbf{CUB} 
and \textbf{SUN} datasets, respectively. 
The activation used was ReLU. 
The image features used were the ResNet-101. 
We compared different variations of our proposed method with previous approaches. 
OURS-R variation is with the training stage 
including the structural loss $\mathcal{L}_2$. 
OURS-RA includes the structural loss as well as the domain adaptation stage 
including the loss functions $\mathcal{L}_3$ and $\mathcal{L}_4$. 
OURS-RC includes the structural loss as well as the calibrated testing stage. 
OURS-RAC includes all the components of structural loss, 
domain adaptation and calibrated testing. 
Without all these components,  the proposed method reduces to 
the Deep Embedding Model (DEM)~\cite{Zhang_2017_CVPR} baseline. 
The parameters $(\lambda_r, \rho, \lambda_g, \gamma)$ 
for the \textbf{AwA2}, \textbf{aPY}, \textbf{CUB} and \textbf{SUN} datasets 
are set as $(10^{-3}, 10^{-1}, 10^{-1}, 1.1)$, 
$(10^{-4}, 10^{-1}, 10^{-1}, 1.1)$, $(10^{-2}, 0, 10^{-1}, 1.1)$ 
and $(10^{-5}, 10^{-1}, 10^{-1}, 1.1)$, respectively. 
For the OURS-RAC variation, 
we used different calibration parameter values of $0.98, 1.1,0.97, 0.999$ 
for the \textbf{AwA2}, \textbf{aPY}, \textbf{CUB} 
and \textbf{SUN} datasets, respectively. 
$\rho$ was set to $0$ for the \textbf{CUB} dataset 
because it is a fine-grained dataset and since the categories 
are very close to each other in the feature space, 
structural matching does not provide additional information. 
In Table~\ref{table:compare}, 
we reported class-wise accuracy results 
for the conventional unseen classes setting (\textbf{tr}), 
generalized unseen classes setting (\textbf{u}), 
generalized seen classes setting (\textbf{s}), 
and the Harmonic mean (\textbf{H}) of the generalized accuracies.

\begin{table*}[]
\caption{Results of variations of our proposed approach in comparison 
with previous methods on the \textbf{AwA2}, \textbf{aPY}, \textbf{CUB} 
and \textbf{SUN} datasets. 
The best results of each setting in each dataset are shown in boldface.}
\label{table:compare}
\centering
\scalebox{1.0}{\begin{tabular}{l|c|c|c|c|c|c|c|c|c|c|c|c|c|c|c|c}
\hline
       & \multicolumn{4}{c|}{\textbf{AwA2}} & \multicolumn{4}{c|}{\textbf{aPY}} & \multicolumn{4}{c|}{\textbf{CUB}} & \multicolumn{4}{c}{\textbf{SUN}} \\ \hline
Method & \textbf{tr}    & \textbf{u}    & \textbf{s}    & \textbf{H}  
& \textbf{tr}    & \textbf{u}    & \textbf{s}    & \textbf{H}  
& \textbf{tr}    & \textbf{u}    & \textbf{s}    & \textbf{H}  
& \textbf{tr}    & \textbf{u}    & \textbf{s}    & \textbf{H}  
\\ \hline
DAP~\cite{lampert2014attribute}    &46.1       &0.0      &84.7      &0.0     &33.8       &4.8      &78.3     &9.0     &40.0       &1.7      &67.9     &3.3     &39.9       &4.2      &25.1     &7.2     \\ 
IAP~\cite{lampert2014attribute}    &35.9       &0.9      &87.6      &1.8     &36.6       &5.7      &65.6     &10.4     &24.0       &0.2      &\textbf{72.8}     &0.4     &19.4       &1.0      &37.8     &1.8     \\ 
CONSE~\cite{norouzi2013zero}   &44.5      &0.5      &\textbf{90.6}      &1.0     &26.9       &0.0      &\textbf{91.2}     &0.0     &34.3       &1.6      &72.2     &3.1     &38.8 &6.8      &39.9     &11.6     \\ 
CMT~\cite{socher2013zero}  &37.9         &0.5      &90.0      &1.0     &28.0       &1.4      &85.2     &2.8     &34.6       &7.2      &49.8     &12.6     &39.9 &8.1      &21.8     &11.8     \\ 
SSE~\cite{zhang2015zero}   &61.0       &8.1      &82.5      &14.8     &34.0       &0.2      &78.9     &0.4     &43.9       &8.5      &46.9     &14.4     &51.5 &2.1      &36.4     &4.0     \\ 
LATEM~\cite{xian2016latent}   &55.8     &11.5      &77.3     &20.0     &35.2       &0.1      &73.0     &0.2     &49.3       &15.2      &57.3     &24.0     &55.3 &14.7      &28.8     &19.5     \\ 
ALE~\cite{akata2016label}    &62.5       &14.0     &81.8     &23.9     &39.7       &4.6      &73.7     &8.7     &54.9       &23.7      &62.8     &34.4     &58.1 &21.8      &33.1     &26.3     \\ 
DEVISE~\cite{frome2013devise}    &59.7    &17.1      &74.7    &27.8     &\textbf{39.8}      &4.9      &76.9     &9.2     &52.0       &23.8      &53.0     &32.8     &56.5 &16.9      &27.4     &20.9     \\ 
SJE~\cite{akata2015evaluation}    &61.9       &8.0      &73.9      &14.4     &32.9       &3.7      &55.7     &6.9     &53.9       &23.5      &59.2     &33.6     &53.7 &14.7      &30.5     &19.8     \\ 
ESZSL~\cite{romera2015embarrassingly}    &58.6       &5.9      &77.8      &11.0     &38.3       &2.4      &70.1  &4.6     &53.9       &12.6      &63.8     &21.0     &54.5 &11.0      &27.9     &15.8     \\ 
SYNC~\cite{changpinyo2016synthesized}   &46.6       &10.0      &90.5      &18.0     &23.9       &7.4      &66.3     &13.3     &55.6       &11.5      &70.9     &19.8     &56.3 &7.9      &\textbf{43.3}     &13.4     \\ 
SAE~\cite{kodirov2017semantic}   &54.1       &1.1      &82.2      &2.2     &8.3       &0.4      &80.9     &0.9     &33.3       &7.8      &54.0     &13.6     &40.3 &8.8      &18.0     &11.8     \\ 
GFZSL~\cite{verma2017simple}   &63.8       &2.5      &80.1     &4.8     &38.4       &0.0      &83.3     &0.0     &49.3       &0.0      &45.7     &0.0     &60.6 &0.0      &39.6     &0.0     \\ 
SR~\cite{annadani2018preserving}   &63.8       &20.7      &73.8      &32.3     &38.4       &13.5      &51.4     &21.4     &\textbf{56.0}       &24.6      &54.3     &33.9     &61.4 &20.8      &37.2     &26.7     \\ 
DEM~\cite{Zhang_2017_CVPR}   &\textbf{67.1}       &30.5      &86.4     &45.1     &35.0       &11.1      &75.1     &19.4     &51.7       &19.6      &57.9     &29.2     &40.3       &20.5      &34.3     &25.6     \\ 
\hline
OURS-R   &63.4    &36.5 &80.6            &50.3     &29.9   &15.3 &71.4          &25.2   &46.6       &20.2 &48.6           &28.6     &59.9      &21.7  &{38.1}          &27.6     \\ 
OURS-RA   &64.4      &\textbf{61.8}   &69.9         &65.6     &35.4        &30.4 &72.9          &42.9     &52.6      &\textbf{47.6}     &41.0       &44.1     &\textbf{67.5}      &\textbf{54.4}  &36.6           &\textbf{43.7}     \\ 
OURS-RC  &63.4      &57.9  &72.0            &64.2     &29.9       &26.4 &53.3    &35.3     &46.6       &27.2      &43.9     &33.6     &59.9        &42.4 &32.6          &36.8     \\ 
OURS-RAC  &64.4            &60.6  &72.3     &\textbf{65.9}     &35.4        &\textbf{34.1}  &63.5         &\textbf{44.4}     &52.6       &44.0  &45.1         &\textbf{44.6}     &\textbf{67.5}      &54.1 &36.6           &\textbf{43.7}     \\ \hline
\end{tabular}}
\vspace*{-0.2in}
\end{table*}

From the table, we observed that our proposed approach 
outperforms previous methods 
by a large margin in the generalized harmonic mean setting. 
To be more specific, our proposed method produces 
an improvement of around 20\%, 23\%, 10\% and 16\% 
harmonic mean accuracy over the previous best approach for the \textbf{AwA2}, \textbf{aPY}, \textbf{CUB} and \textbf{SUN} datasets, respectively. 
The large improvement in performance can be attributed 
to our three-step procedure for improvement. 
Using only the structural matching (OURS-R), 
we produced better results than previous approaches 
except for the \textbf{CUB} dataset, 
where it produces a harmonic mean accuracy of about 28\%. 
This is because \textbf{CUB} requires minute fine-grained feature extraction. 
Additional usage of domain adaptation (OURS-RA) 
and calibrated testing (OURS-RC) produced 
much better results than OURS-R for all the datasets. 
However, domain adaptation produced better result than the calibration procedure. 
This is because our correspondence-based approach 
produced class-specific adaptation of the unseen class semantic embeddings. 
The scaled-calibration procedure is not class-specific 
and just differentiates between seen and unseen classes. 
It also does not adapt to the test data. 

\begin{figure}[]
\centering
\includegraphics[width=8cm]{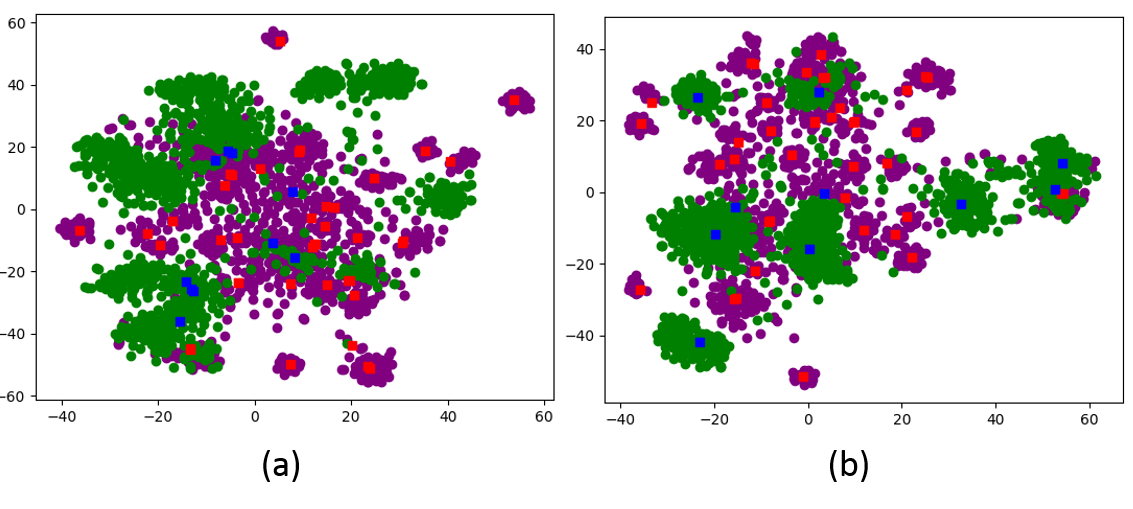}
\vspace*{-0.1in}
\caption{2D t-SNE map of the embedded instances. (a) Without domain adaptation and 
(b) with domain adaptation for the \textbf{AwA2} dataset. 
Here, the seen and unseen image features are shown in maroon and green, respectively. 
The embedded semantic descriptors for the seen and unseen classes 
are shown in red and blue color, respectively. }
\label{fig:tsnezsl1}
\vspace*{-0.2in}
\end{figure}

It is to be noted that the difference in performance 
between OURS-RA and OURS-RAC is negligible. 
This is because the domain adaptation step transforms 
the unseen semantic embeddings away 
from the seen categories towards the unseen categories, 
thus reducing the bias towards the seen categories 
and rendering further calibration ineffective. 
The effect of domain adaptation is visualized in 
Fig.~\ref{fig:tsnezsl1} for the \textbf{AwA2} dataset 
using t-SNE~\cite{maaten2008visualizing}. 
In Fig.~\ref{fig:tsnezsl1}(a), 
the unseen class semantic embeddings (blue) remained very close to 
the seen class features (maroon). 
However, with the domain adaptation step, 
the unseen class semantic embeddings get transformed to 
near the centre of unseen class feature clusters (green) 
as shown in Fig.~\ref{fig:tsnezsl1}(b).

\begin{figure}[]
\centering
\includegraphics[width=8cm]{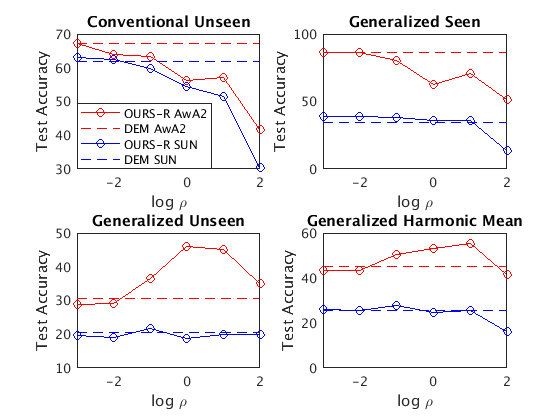}
\vspace*{-0.1in}
\caption{Results of class-wise accuracy as $\rho$ is varied for different settings 
on the \textbf{AwA2} and the \textbf{SUN} datasets. 
The baseline used is DEM. The different performance settings 
are Conventional Unseen Accuracy (Left Top), 
Generalized Seen Accuracy (Right Top), Generalized Unseen Accuracy (Left Bottom) 
and Generalized Harmonic Mean Accuracy (Right Bottom).}
\label{fig:rhosens}
\vspace*{-0.2in}
\end{figure}

We also analyzed the effect of the structural matching 
by varying $\rho \in \{10^{-3}, 10^{-2}, 10^{-1}, 10^{0}, 10^{1}, 10^{2}\}$ 
and observed how the class-wise accuracy changes. 
We carried out experiments using the \textbf{AwA2} and \textbf{SUN} datasets, 
the results of which are reported in Fig.~\ref{fig:rhosens}. 
We also reported the DEM baseline ($\rho=0$) in dotted lines. 
From the plots, the Conventional Unseen and the Generalized Seen accuracies 
are better than or equal to the baseline for only a small range of $\rho$. 
On the other hand, the Generalized Unseen accuracy is greater than the baseline 
over a large range of $\rho$ for the \textbf{AwA2} dataset 
while it oscillated about the baseline for the \textbf{SUN} dataset. 
For the \textbf{SUN} dataset, 
we do not have a significant gain over the baseline 
because \textbf{SUN} is a fine-grained dataset 
where structural matching does not carry additional information. 
The goal of structural regularization is to exploit 
the pairwise relations among classes so as to generalize better to novel classes. 
Therefore, we did not see huge difference in performance 
from the baseline for the Generalized Seen accuracy. 
Surprisingly, there was a drop in conventional unseen accuracy 
as $\rho$ was increased. 
This might be probably because there was no overlap 
between the classes used for testing and the classes used for structural matching. 
This is not the case though in the generalized setting. 

\begin{figure}[]
\centering
\includegraphics[width=8cm]{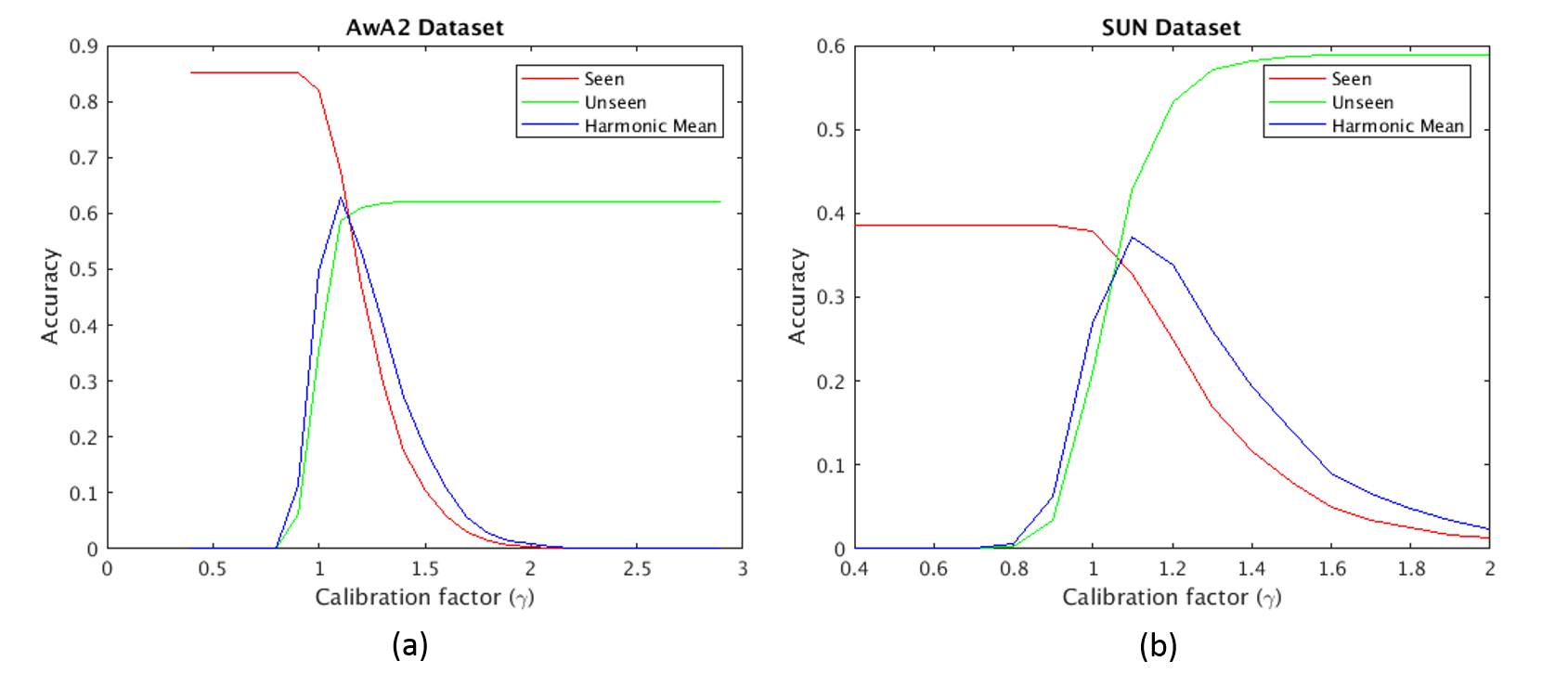}
\vspace*{-0.1in}
\caption{Results of Generalized Seen Accuracy (Red), 
Generalized Unseen Accuracy (Green) 
and Generalized Harmonic Mean Accuracy (Blue) 
as the calibration parameter $\gamma$ is varied 
on the \textbf{AwA2} and \textbf{SUN} datasets.}
\label{fig:calib}
\vspace*{-0.2in}
\end{figure}

We also studied the effect of varying the calibration parameter $\gamma$ 
on the generalized accuracy for the \textbf{AwA2} and \textbf{SUN} datasets. 
The results are shown in Fig.~\ref{fig:calib}. 
As expected, the generalized unseen accuracy increases 
and the generalized seen accuracy decreases with increasing $\gamma$. 
The peak of the harmonic mean accuracy was observed close to 
when the seen and unseen accuracies became equal. 
The maximum unseen accuracy is less than the maximum seen accuracy 
for the \textbf{AwA2} dataset because the unseen classes 
are less separated and therefore more difficult to classify. 
The situation is reversed for the \textbf{SUN} dataset 
where the maximum unseen accuracy is more than the maximum seen accuracy.

\begin{figure}[]
\centering
\includegraphics[width=8cm]{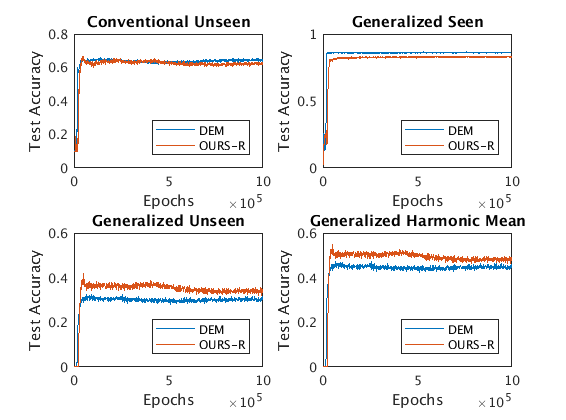}
\vspace*{-0.1in}
\caption{Convergence results of test accuracy with respect to the number of epochs 
under different settings for the \textbf{AwA2} dataset. 
OURS-R results are shown in red color while the DEM baseline is shown in blue color.}
\label{fig:epAwA2}
\vspace*{-0.2in}
\end{figure}

\begin{figure}[]
\centering
\includegraphics[width=8cm]{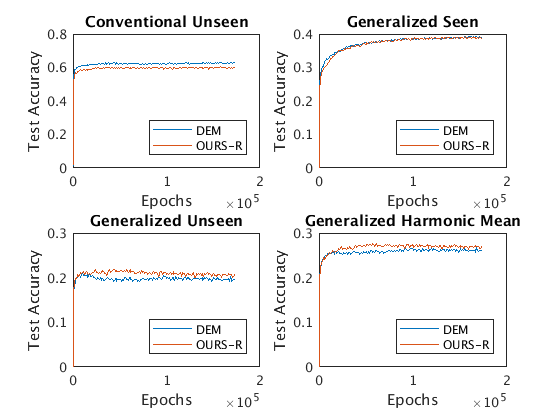}
\vspace*{-0.1in}
\caption{Convergence results of test accuracy 
with respect to the number of epochs under different settings 
for the \textbf{SUN} dataset. 
OURS-R results are shown in red color while the DEM baseline is shown in blue color.}
\label{fig:epSUN}
\vspace*{-0.2in}
\end{figure}

We also reported convergence results of the test accuracy 
with respect to the number of epochs for both the \textbf{AwA2} 
and the \textbf{SUN} datasets in Figs.~\ref{fig:epAwA2} and~\ref{fig:epSUN}, respectively.
We used the OURS-R variation with $\rho=0.1$ to compare with the DEM baseline. 
The convergence rate for the baseline and OURS-R variation seems 
to be similar in all the settings for both datasets. 
However, our steady-state values were higher 
for the generalized unseen and generalized harmonic mean setting. 
For the conventional unseen and generalized seen setting, 
our steady-state value was less than the baseline. 
The reason is explained previously 
while describing performance sensitivity to $\rho$. 

\begin{figure}[]
\centering
\includegraphics[width=6cm]{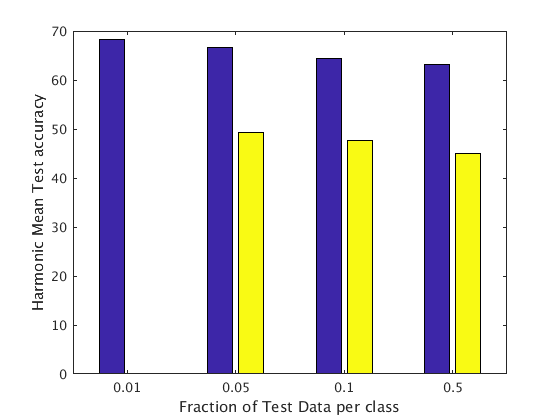}
\vspace*{-0.1in}
\caption{Generalized Harmonic Mean Accuracy results 
as the number of test samples per class is varied for the \textbf{AwA2} (blue) 
and \textbf{SUN} (yellow) datasets.}
\label{fig:nsamp}
\vspace*{-0.2in}
\end{figure}

We also studied the effect of varying the number of test unseen samples per class 
on the generalized harmonic mean accuracy. 
We used OURS-RA variation of our model for this study. 
$\rho=0.1$ was set for the experiments on the \textbf{AwA2} (blue color) 
and the \textbf{SUN} (yellow color) datasets 
and the result was reported in Fig.~\ref{fig:nsamp}. 
When the fraction is 0.01 for the \textbf{SUN} dataset, 
the number of samples in some classes becomes zero 
and therefore the performance is not reported. 
From the results, it is seen that the test accuracy was stable 
with change in the fraction of total number of samples used for testing. 
There is a slight increase in accuracy with decreasing number of samples, 
which is surprising because domain adaptation would perform poorly 
with less number of samples. 
However, this effect is nullified since the probability 
of including challenging examples is reduced 
and so we observed a slight improvement in performance.

\begin{figure}[]
\centering
\includegraphics[width=6cm]{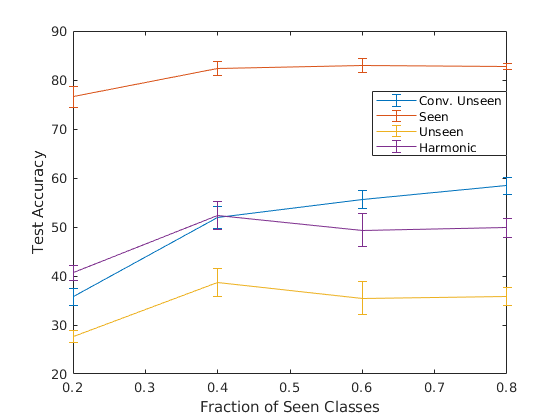}
\vspace*{-0.1in}
\caption{Test accuracy results as the number of seen classes 
used for training is varied for the \textbf{AwA2} dataset.}
\label{fig:nclass}
\vspace*{-0.2in}
\end{figure}

We also studied how the test performance varies 
as the number of seen classes for training is reduced 
for the \textbf{AwA2} dataset using OURS-R model. 
We set $\rho=0.1$ and reported results over 5 trials in Fig.~\ref{fig:nclass}. 
We observed that the change in the seen-class accuracy 
is not much because the training and testing distributions are the same. 
The conventional unseen-class accuracy dips by a large amount 
as the number of training classes decreases because 
there is less representative information to be transferred to novel categories. 
However, we obtained a peak for the generalized unseen accuracy results 
at a fraction of 0.4 of the number of seen classes. 
This is because as the number of training classes decreases, 
the amount of representative information decreases, causing decrease in performance. 
On the other hand, less number of seen classes implies 
less bias towards seen categories and improvement of unseen-class accuracies. 
Also, there is large performance variation for unseen-class accuracy 
because training and testing distributions are different 
and the performance can vary depending on 
how related are the training classes to the unseen classes in a trial.

We also performed experiments to find whether the OURS-R variant 
reduces hubness compared to DEM. 
The hubness of a set of predictions is measured 
using the skewness of the 1-Nearest-Neighbor histogram ($N_1$). 
The $N_1$ histogram is a frequency plot for $N_1[i]$ 
of the number of times a search solution $i$ (in our case a class attribute) 
is found as the Nearest Neighbor for the test samples. 
Less skewness of $N_1$ histogram implies less hubness of the predictions. 
We used the test samples of the unseen classes in the generalized setting 
for both DEM and OURS-R on the \textbf{AwA2} and the \textbf{aPY} datasets. 
We used $\rho=0.1$ and reported results averaging over 5 trials 
in Table~\ref{table:hubness}. 
From the results, 
OURS-R method produced less skewness of the $N_1$ histogram on both the datasets. 
This implies that using the additional structural term reduces 
hubness and therefore the curse of dimensionality is reduced.

\begin{table}[]
\center
\caption{Hubness comparison using skewness for DEM and OURS-R methods 
on the \textbf{AwA2} and \textbf{aPY} datasets}
\label{table:hubness}
\begin{tabular}{@{}ccc@{}}
\toprule
Skewness & AwA2  & aPY \\ \midrule
DEM      & 3.39    & 1.85  \\
OURS-R   & 2.41    & 1.33  \\ \bottomrule
\end{tabular}
\end{table}

\section{Conclusion}
This paper proposed a three-step approach to improve 
the performance of zero-shot learning for image classification. 
The three-step approach involved exploiting structural information in data, 
domain adaptation to unseen test samples and calibration of classification scores. 
When the proposed method was applied to 
standard datasets of zero-shot image classification, 
it outperformed previous methods by a large margin, where
the most effective component was the domain adaptation step. 
\bibliographystyle{IEEEbib}
\bibliography{TLMaster}

\begin{thebibliography}{10}

\bibitem{deng2009imagenet}
Jia Deng, Wei Dong, Richard Socher, Li-Jia Li, Kai Li, and Li~Fei-Fei,
\newblock ``Imagenet: A large-scale hierarchical image database,''
\newblock in {\em Proc. {IEEE} Conf. Comput. Vis. Pattern Recog. (CVPR)}, 2009,
  pp. 248--255.

\bibitem{krizhevsky2012imagenet}
Alex Krizhevsky, Ilya Sutskever, and Geoffrey~E Hinton,
\newblock ``Imagenet classification with deep convolutional neural networks,''
\newblock in {\em Advan. Neu. Inf. Proc. Syst.}, 2012, pp. 1097--1105.

\bibitem{lampert2014attribute}
Christoph~H Lampert, Hannes Nickisch, and Stefan Harmeling,
\newblock ``Attribute-based classification for zero-shot visual object
  categorization,''
\newblock {\em {IEEE} Trans. Pattern Anal. Mach. Intell.}, vol. 36, no. 3, pp.
  453--465, 2014.

\bibitem{word2vec}
Tomas Mikolov, Ilya Sutskever, Kai Chen, Greg~S Corrado, and Jeff Dean,
\newblock ``Distributed representations of words and phrases and their
  compositionality,''
\newblock in {\em Advan. Neu. Inf. Proc. Syst.}, 2013, pp. 3111--3119.

\bibitem{Zhang_2017_CVPR}
Li~Zhang, Tao Xiang, and Shaogang Gong,
\newblock ``Learning a deep embedding model for zero-shot learning,''
\newblock in {\em Proc. {IEEE} Conf. Comput. Vis. Pattern Recog. (CVPR)}, 2017.

\bibitem{akata2016label}
Zeynep Akata, Florent Perronnin, Zaid Harchaoui, and Cordelia Schmid,
\newblock ``Label-embedding for image classification,''
\newblock {\em {IEEE} Trans. Pattern Anal. Mach. Intell.}, vol. 38, no. 7, pp.
  1425--1438, 2016.

\bibitem{frome2013devise}
Andrea Frome, Greg~S Corrado, Jon Shlens, Samy Bengio, Jeff Dean, Tomas
  Mikolov, et~al.,
\newblock ``Devise: A deep visual-semantic embedding model,''
\newblock in {\em Advan. Neu. Inf. Proc. Syst.}, 2013, pp. 2121--2129.

\bibitem{socher2013zero}
Richard Socher, Milind Ganjoo, Christopher~D Manning, and Andrew Ng,
\newblock ``Zero-shot learning through cross-modal transfer,''
\newblock in {\em Advan. Neu. Inf. Proc. Syst.}, 2013, pp. 935--943.

\bibitem{zhang2016zero}
Ziming Zhang and Venkatesh Saligrama,
\newblock ``Zero-shot learning via joint latent similarity embedding,''
\newblock in {\em Proc. {IEEE} Conf. Comput. Vis. Pattern Recog. (CVPR)}, 2016,
  pp. 6034--6042.

\bibitem{changpinyo2016synthesized}
Soravit Changpinyo, Wei-Lun Chao, Boqing Gong, and Fei Sha,
\newblock ``Synthesized classifiers for zero-shot learning,''
\newblock in {\em Proc. {IEEE} Conf. Comput. Vis. Pattern Recog. (CVPR)}, 2016,
  pp. 5327--5336.

\bibitem{romera2015embarrassingly}
Bernardino Romera-Paredes and Philip Torr,
\newblock ``An embarrassingly simple approach to zero-shot learning,''
\newblock in {\em Intern. Conf. Mach. Learn.}, 2015, pp. 2152--2161.

\bibitem{akata2015evaluation}
Zeynep Akata, Scott Reed, Daniel Walter, Honglak Lee, and Bernt Schiele,
\newblock ``Evaluation of output embeddings for fine-grained image
  classification,''
\newblock in {\em Proc. {IEEE} Conf. Comput. Vis. Pattern Recog. (CVPR)}, 2015,
  pp. 2927--2936.

\bibitem{xian2016latent}
Yongqin Xian, Zeynep Akata, Gaurav Sharma, Quynh Nguyen, Matthias Hein, and
  Bernt Schiele,
\newblock ``Latent embeddings for zero-shot classification,''
\newblock in {\em Proc. {IEEE} Conf. Comput. Vis. Pattern Recog. (CVPR)}, 2016,
  pp. 69--77.

\bibitem{kodirov2017semantic}
Elyor Kodirov, Tao Xiang, and Shaogang Gong,
\newblock ``Semantic autoencoder for zero-shot learning,''
\newblock in {\em Proc. {IEEE} Conf. Comput. Vis. Pattern Recog. (CVPR)}, 2017,
  pp. 4447--4456.

\bibitem{radovanovic2010hubs}
Milo{\v{s}} Radovanovi{\'c}, Alexandros Nanopoulos, and Mirjana Ivanovi{\'c},
\newblock ``Hubs in space: Popular nearest neighbors in high-dimensional
  data,''
\newblock {\em Journal of Machine Learning Research}, vol. 11, no. Sep, pp.
  2487--2531, 2010.

\bibitem{shigeto2015ridge}
Yutaro Shigeto, Ikumi Suzuki, Kazuo Hara, Masashi Shimbo, and Yuji Matsumoto,
\newblock ``Ridge regression, hubness, and zero-shot learning,''
\newblock in {\em Joint European Conference on Machine Learning and Knowledge
  Discovery in Databases}, 2015, pp. 135--151.

\bibitem{2stage}
Debasmit Das and C.~S.~George Lee,
\newblock ``A two-stage approach to few-shot learning for image recognition,''
\newblock {\em Working Paper}, 2019.

\bibitem{fu2015transductive}
Yanwei Fu, Timothy~M Hospedales, Tao Xiang, and Shaogang Gong,
\newblock ``Transductive multi-view zero-shot learning,''
\newblock {\em {IEEE} Trans. Pattern Anal. Mach. Intell.}, vol. 37, no. 11, pp.
  2332--2345, 2015.

\bibitem{kodirov2015unsupervised}
Elyor Kodirov, Tao Xiang, Zhenyong Fu, and Shaogang Gong,
\newblock ``Unsupervised domain adaptation for zero-shot learning,''
\newblock in {\em Proc. {IEEE} Int. Conf. Comput. Vis.}, 2015, pp. 2452--2460.

\bibitem{das2018sample}
Debasmit Das and C.~S.~George Lee,
\newblock ``Sample-to-sample correspondence for unsupervised domain
  adaptation,''
\newblock {\em Engineering Applications of Artificial Intelligence}, vol. 73,
  pp. 80--91, 2018.

\bibitem{chao2016empirical}
Wei-Lun Chao, Soravit Changpinyo, Boqing Gong, and Fei Sha,
\newblock ``An empirical study and analysis of generalized zero-shot learning
  for object recognition in the wild,''
\newblock in {\em European Conference on Computer Vision}, 2016, pp. 52--68.

\bibitem{zhang2015zero}
Ziming Zhang and Venkatesh Saligrama,
\newblock ``Zero-shot learning via semantic similarity embedding,''
\newblock in {\em Proc. {IEEE} Int. Conf. Comput. Vis.}, 2015, pp. 4166--4174.

\bibitem{norouzi2013zero}
Mohammad Norouzi, Tomas Mikolov, Samy Bengio, Yoram Singer, Jonathon Shlens,
  Andrea Frome, Greg~S Corrado, and Jeffrey Dean,
\newblock ``Zero-shot learning by convex combination of semantic embeddings,''
\newblock in {\em Intern. Conf. Learning Representations}, 2013.

\bibitem{verma2018generalized}
Vinay~Kumar Verma, Gundeep Arora, Ashish Mishra, and Piyush Rai,
\newblock ``Generalized zero-shot learning via synthesized examples,''
\newblock in {\em Proc. {IEEE} Conf. Comput. Vis. Pattern Recog. (CVPR)}, 2018.

\bibitem{guo2017synthesizing}
Yuchen Guo, Guiguang Ding, Jungong Han, and Yue Gao,
\newblock ``Synthesizing samples for zero-shot learning,''
\newblock in {\em Proceedings of the 26th International Joint Conference on
  Artificial Intelligence}. AAAI Press, 2017, pp. 1774--1780.

\bibitem{xian2018zero}
Yongqin Xian, Christoph~H Lampert, Bernt Schiele, and Zeynep Akata,
\newblock ``Zero-shot learning-a comprehensive evaluation of the good, the bad
  and the ugly,''
\newblock {\em {IEEE} Trans. Pattern Anal. Mach. Intell.}, 2018.

\bibitem{frank1956algorithm}
Marguerite Frank and Philip Wolfe,
\newblock ``An algorithm for quadratic programming,''
\newblock {\em Naval Research Logistics (NRL)}, vol. 3, no. 1-2, pp. 95--110,
  1956.

\bibitem{bonneel2011displacement}
Nicolas Bonneel, Michiel Van De~Panne, Sylvain Paris, and Wolfgang Heidrich,
\newblock ``Displacement interpolation using lagrangian mass transport,''
\newblock {\em ACM Trans. Graphics (TOG)}, vol. 30, no. 6, pp. 158, 2011.

\bibitem{farhadi2009describing}
Ali Farhadi, Ian Endres, Derek Hoiem, and David Forsyth,
\newblock ``Describing objects by their attributes,''
\newblock in {\em Proc. {IEEE} Conf. Comput. Vis. Pattern Recog. (CVPR)}, 2009,
  pp. 1778--1785.

\bibitem{welinder2010caltech}
P.~Welinder, S.~Branson, T.~Mita, C.~Wah, F.~Schroff, S.~Belongie, and
  P.~Perona,
\newblock ``{Caltech-UCSD Birds 200},''
\newblock Tech. {R}ep. CNS-TR-2010-001, California Institute of Technology,
  2010.

\bibitem{patterson2012sun}
Genevieve Patterson and James Hays,
\newblock ``Sun attribute database: Discovering, annotating, and recognizing
  scene attributes,''
\newblock in {\em Proc. {IEEE} Conf. Comput. Vis. Pattern Recog. (CVPR)}, 2012,
  pp. 2751--2758.

\bibitem{verma2017simple}
Vinay~Kumar Verma and Piyush Rai,
\newblock ``A simple exponential family framework for zero-shot learning,''
\newblock in {\em Joint European Conference on Machine Learning and Knowledge
  Discovery in Databases}, 2017, pp. 792--808.

\bibitem{annadani2018preserving}
Yashas Annadani and Soma Biswas,
\newblock ``Preserving semantic relations for zero-shot learning,''
\newblock in {\em Proc. {IEEE} Conf. Comput. Vis. Pattern Recog. (CVPR)}, 2018,
  pp. 7603--7612.

\bibitem{maaten2008visualizing}
Laurens van~der Maaten and Geoffrey Hinton,
\newblock ``Visualizing data using t-sne,''
\newblock {\em Journal of Machine Learning Research}, vol. 9, no. Nov, pp.
  2579--2605, 2008.

\end{thebibliography}

%
%

%

\begin{IEEEbiography}{Debasmit Das}
Biography text here.
\end{IEEEbiography}

\begin{IEEEbiography}{C.S. George Lee}
Biography text here.
\end{IEEEbiography}






\end{document}